\documentclass{vgtc}                          

\graphicspath{{figures/}{pictures/}{images/}{./}} 

\usepackage{times}                     

\usepackage{tabu}                      
\usepackage{booktabs}                  
\usepackage{lipsum}                    
\usepackage{mwe}                       

\usepackage{mathptmx}                  
\usepackage{amsmath}                   

\onlineid{0}

\vgtccategory{Research}

\vgtcinsertpkg

\title{GazeProphetV2: Head-Movement-Based Gaze Prediction Enabling Efficient Foveated Rendering on Mobile VR}

\author{Farhaan Ebadulla\thanks{e-mail: PES2202101091@pesu.pes.edu} %
\and Chiraag Mudlapur\thanks{e-mail:mudlapur.chiraag@gmail.com} %
\and Shreya Chaurasia\thanks{e-mail:shreyapjchaurasia04@gmail.com} %
\and Gaurav BV\thanks{e-mail:bv.gaurav4@gmail.com}}
\affiliation{\scriptsize PES University}

\abstract{
    Predicting gaze behavior in virtual reality environments remains a significant challenge with implications for rendering optimization and interface design. This paper introduces a multimodal approach to VR gaze prediction that combines temporal gaze patterns, head movement data, and visual scene information. By leveraging a gated fusion mechanism with cross-modal attention, the approach learns to adaptively weight gaze history, head movement, and scene content based on contextual relevance. Evaluations using a dataset spanning 22 VR scenes with 5.3M gaze samples demonstrate improvements in predictive accuracy when combining modalities compared to using individual data streams alone. The results indicate that integrating past gaze trajectories with head orientation and scene content enhances prediction accuracy across 1-3 future frames. Cross-scene generalization testing shows consistent performance with 93.1\% validation accuracy and temporal consistency in predicted gaze trajectories. These findings contribute to understanding attention mechanisms in virtual environments while suggesting potential applications in rendering optimization, interaction design, and user experience evaluation. The approach represents a step toward more efficient virtual reality systems that can anticipate user attention patterns without requiring expensive eye tracking hardware.
} 


\begin{document}



\firstsection{Introduction}
\maketitle
Virtual reality applications demand substantial computational resources to deliver immersive experiences that meet modern visual quality standards. Contemporary VR systems must render high-resolution content at sustained frame rates exceeding 90 Hz while maintaining low latency to prevent motion sickness~\cite{patney2016towards}. This computational burden becomes increasingly challenging as VR displays advance toward higher resolutions and wider fields of view, creating a fundamental bottleneck for widespread VR adoption.

Foveated rendering presents a promising solution to these computational constraints by exploiting the natural characteristics of human vision. The technique selectively reduces rendering quality in peripheral regions while maintaining high detail in the central foveal area where visual acuity concentrates~\cite{singh2023power}. Studies demonstrate that properly implemented foveated rendering can achieve computational savings of up to 70\% without perceptible quality degradation~\cite{kaplanyan2019deepfovea}. This approach aligns with biological vision systems, where only the central 2 degrees of the visual field provide maximum acuity.

Current foveated rendering implementations rely heavily on dedicated eye tracking hardware to determine gaze locations in real-time. Commercial VR systems incorporating eye tracking, such as the Meta Quest Pro~\cite{meta2022questpro} and HTC Vive Pro Eye~\cite{htc2019vive}, utilize infrared cameras and specialized sensors to monitor eye movements~\cite{adhanom2023eye}. However, these systems face significant practical limitations. Eye tracking hardware adds substantial cost to VR headsets, requires precise calibration procedures, and exhibits accuracy constraints that impact foveated rendering effectiveness~\cite{adhanom2020gazemetrics}.

Recent evaluations reveal that current VR eye tracking systems achieve accuracy between 1.2-1.7 degrees, falling short of manufacturer specifications claiming 0.5-1.1 degree precision~\cite{wei2023preliminary}. These accuracy limitations directly affect foveated rendering performance, as incorrect gaze estimates can produce visible artifacts or inappropriate quality reduction. Additionally, eye tracking hardware remains absent from the majority of consumer VR devices, limiting foveated rendering benefits to premium systems.

Software-based gaze prediction offers an alternative approach that eliminates hardware dependencies while potentially achieving comparable performance. Recent advances in computer vision and deep learning enable sophisticated models to predict gaze locations using scene content and behavioral patterns~\cite{xu2018gaze}. Multimodal approaches that integrate multiple information sources demonstrate particular promise, showing 10-21\% improvement over single-modality techniques~\cite{gupta2022modular}.

The temporal nature of gaze behavior provides valuable predictive information often underexplored in existing approaches. Gaze patterns exhibit strong sequential dependencies, with users following predictable scan paths and returning to previously viewed regions. Head movement data offers additional predictive signals, as gaze shifts in VR environments typically follow coordinated head-eye movement patterns~\cite{sidenmark2019eye}. Research demonstrates that head movements often precede gaze shifts by 200-400 milliseconds, creating opportunities for predictive modeling.

This paper presents a multimodal approach for VR gaze prediction that combines temporal gaze patterns, head movement data, and visual scene information. The approach employs an LSTM-based temporal encoder to capture sequential gaze dependencies while incorporating head orientation through quaternion-based features. A fusion mechanism adaptively weights different modalities based on confidence estimates and temporal context. Multi-step prediction enables anticipation of future gaze locations across multiple timesteps, supporting adaptive foveated rendering strategies.

The contributions include: 
\begin{itemize}
    \item a multimodal fusion architecture that integrates gaze history, head movement, and scene content for enhanced prediction accuracy,
    \item temporal modeling through LSTM networks that captures sequential dependencies in gaze behavior,
    \item multi-step prediction capability enabling anticipation of future gaze locations,
    \item comprehensive evaluation demonstrating improvements over baseline approaches on VR-specific datasets.
\end{itemize}

The findings demonstrate that software-based gaze prediction represents a viable solution for VR rendering optimization. This approach facilitates foveated rendering implementation on current VR hardware without requiring additional equipment investments. Such accessibility expands performance enhancement opportunities across a broader spectrum of VR systems and application domains, removing traditional barriers to advanced rendering techniques.

\section{Related Work}

This section reviews relevant research across six key areas that inform multimodal gaze prediction for VR systems. The literature demonstrates both the necessity for software-based approaches and the potential benefits of integrating multiple information modalities.

\subsection{Eye Tracking in VR Systems}

Eye tracking technology in virtual reality has evolved significantly over the past decade, yet fundamental limitations persist. Commercial systems like the Tobii Pro VR Integration~\cite{tobii2023vr} and SMI Eye Tracking HMD~\cite{smi2022hmd} provide high-precision gaze tracking for research applications. These systems achieve sub-degree accuracy under controlled conditions. However, they require specialized hardware integration that substantially increases system complexity and cost.

Consumer VR headsets have begun incorporating eye tracking capabilities. The HTC Vive Pro Eye~\cite{htc2019vive} introduced eye tracking to mainstream VR gaming platforms. Meta Quest Pro~\cite{meta2022questpro} includes advanced eye tracking sensors for social applications. Apple Vision Pro~\cite{apple2023vision} employs sophisticated eye tracking for user interface control. PlayStation VR2~\cite{sony2023psvr2} integrates eye tracking with native foveated rendering support.

Despite these advances, eye tracking remains expensive and technically challenging. Comprehensive evaluations reveal significant gaps between manufacturer specifications and real-world performance. Adhanom et al.~\cite{adhanom2023eye} demonstrate that current VR eye tracking systems achieve accuracy between 1.2-1.7 degrees. This substantially exceeds claimed precision of 0.5-1.1 degrees. Hardware costs add hundreds of dollars to headset prices. Calibration procedures require user training and frequent recalibration~\cite{adhanom2020gazemetrics}. Individual differences in eye anatomy affect tracking accuracy. Environmental factors like lighting conditions impact system performance~\cite{wei2023preliminary}.

\subsection{Foveated Rendering Approaches}

Foveated rendering techniques aim to reduce computational load while maintaining visual quality. These approaches exploit human visual system characteristics. Early implementations used fixed foveated rendering~\cite{murphy2009gaze}. They applied reduced quality to predefined peripheral regions without tracking actual gaze locations. These approaches provided limited benefits due to their static nature.

Dynamic foveated rendering adapts quality based on real-time gaze tracking~\cite{patney2016towards,guenter2012foveated}. Research demonstrates significant performance improvements with minimal perceptual impact. Patney et al.~\cite{patney2016towards} show 50-70\% rendering cost reductions while maintaining visual fidelity. Recent research explores perceptually-guided approaches that incorporate detailed human visual system models~\cite{kaplanyan2019deepfovea}.

Temporal aspects of vision receive attention through motion-aware foveated rendering~\cite{albert2017latency}. Multi-resolution approaches balance quality and performance across different viewing conditions~\cite{deng2022fovnerf}. Singh et al.~\cite{singh2023power} provide comprehensive analysis of power, performance, and image quality tradeoffs. They demonstrate up to 70\% computational savings with properly calibrated systems. However, these benefits require accurate gaze prediction to avoid visible artifacts from incorrect quality allocation.

\subsection{Gaze Prediction Approaches}

Visual attention prediction has extensive research history in computer vision. The field has evolved from traditional saliency models to sophisticated deep learning approaches. Early methods used hand-crafted features to identify attention-grabbing regions~\cite{itti1998saliency,harel2007graph}. These approaches focused on low-level visual features like color, intensity, and orientation contrast.

Deep learning transformed saliency prediction capabilities. Convolutional neural networks demonstrate superior performance over traditional methods~\cite{kruthiventi2017deepfix,cornia2018lstm}. Attention mechanisms enhance model interpretability and performance~\cite{wang2018deep}. Video saliency prediction incorporates temporal information for dynamic scenes~\cite{mathe2016actions,gorji2019video}. These approaches use recurrent neural networks and LSTM architectures for sequence modeling~\cite{linardos2021simple}.

Gaze prediction in 360-degree content presents unique challenges due to spherical image geometry~\cite{xu2018gaze,chao2018salnet360}. VR environments require temporal modeling of gaze sequences for effective prediction~\cite{david2018dataset}. Recent approaches address these challenges through spherical convolutions and specialized attention mechanisms~\cite{ling2022saltivr}. However, most existing approaches focus on single-modality prediction, leaving opportunities for multimodal integration unexplored.

Previous work~\cite{ebadulla2025gazeprophet} demonstrated baseline comparisons between traditional saliency methods and deep learning approaches for VR environments, establishing the foundation for multimodal integration explored in this work.

\subsection{Multimodal Fusion Techniques}

Multimodal approaches for gaze prediction demonstrate superior performance over single-modality techniques. Gupta et al.~\cite{gupta2022modular} present a modular architecture that achieves 10-21\% improvement over single-stream approaches. Their approach integrates visual scene content with behavioral patterns. The architecture employs separate encoders for different modalities with learned fusion weights.

Nonaka et al.~\cite{nonaka2022dynamic} explore temporal eye-head-body coordination for dynamic 3D gaze estimation. They demonstrate that multi-modal temporal information significantly improves prediction accuracy in unconstrained settings. Tafasca et al.~\cite{tafasca2024sharingan} introduce transformer-based architectures for multi-person gaze following. Their research shows that attention mechanisms effectively integrate spatial and temporal modalities.

Fusion strategies vary from early concatenation to sophisticated attention-based approaches. Baltrusaitis et al.~\cite{baltrusaitis2019multimodal} provide comprehensive taxonomy of multimodal machine learning techniques. They highlight the importance of learned fusion weights over static combination strategies. Zadeh et al.~\cite{zadeh2017tensor} demonstrate that tensor fusion networks enable complex multimodal interactions beyond simple concatenation approaches.

\subsection{Temporal Modeling for Gaze Prediction}

Long Short-Term Memory networks excel at capturing temporal dependencies in sequential gaze data~\cite{hochreiter1997lstm}. LSTM architectures address vanishing gradient problems that affect standard recurrent networks in long sequence modeling. Applications in computer vision include video analysis~\cite{donahue2015long} and sequential prediction tasks~\cite{xingjian2015convolutional}.

Gaze sequence modeling benefits significantly from temporal architectures. Previous gaze locations provide strong predictive signals for future attention patterns~\cite{zelinsky2013understanding}. Temporal patterns vary across individuals and tasks, requiring adaptive modeling approaches~\cite{henderson2003human}. Recent research combines convolutional and recurrent architectures for improved spatiotemporal modeling~\cite{huang2015salicon}.

Transformer architectures show promise for temporal gaze modeling. Recent approaches demonstrate that self-attention mechanisms effectively capture long-range dependencies in gaze sequences~\cite{park2018deep,park2019few}. However, most temporal approaches focus on single-modality inputs, missing opportunities to incorporate multimodal temporal relationships.

\subsection{Head-Gaze Coordination Research}

\begin{figure*}[htbp]
\centering
\includegraphics[width=0.9\textwidth]{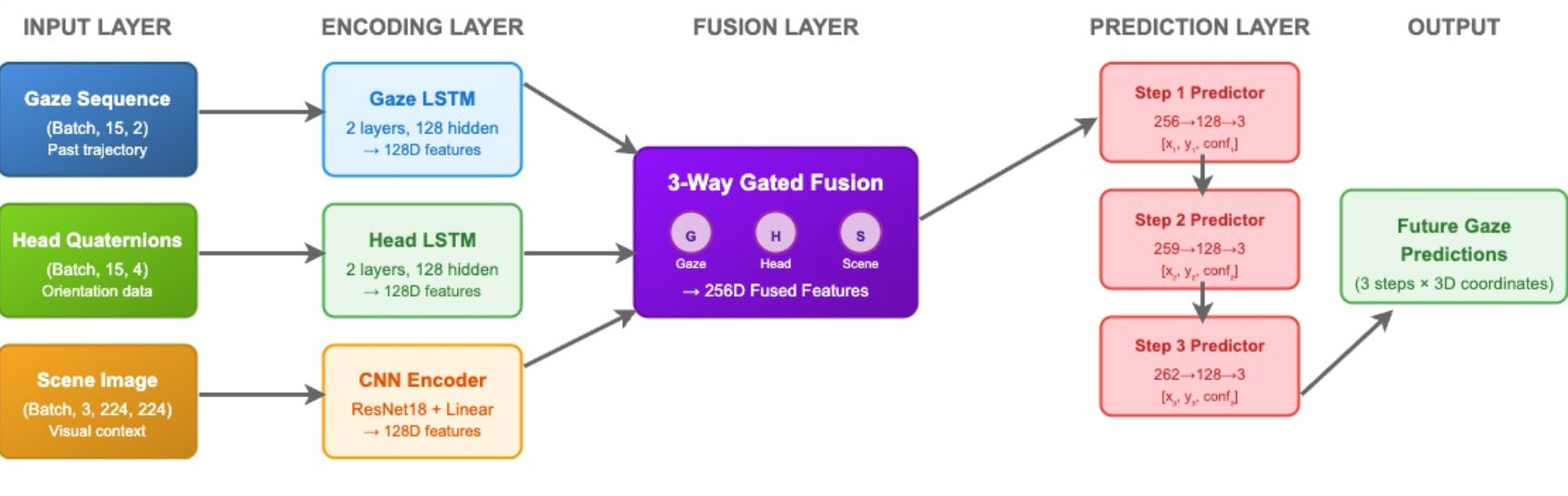}
\caption{Complete system architecture showing CNN scene encoder, LSTM temporal encoders for gaze and head sequences, three-way gated fusion mechanism, and multi-step prediction heads. The architecture processes 15-frame input sequences to generate 1, 2, and 3-frame ahead gaze predictions with auxiliary losses ensuring individual encoder contributions.}
\label{fig:architecture}
\end{figure*}

Research demonstrates strong coordination between head and eye movements in VR environments. Sidenmark and Gellersen~\cite{sidenmark2019eye} provide comprehensive analysis of eye, head, and torso coordination during gaze shifts in virtual reality. Their findings show that head movements typically precede gaze shifts by 200-400 milliseconds. This creates predictive opportunities for gaze estimation systems.

VR environments naturally couple head and eye movements due to immersive interaction paradigms. Users typically orient their heads toward regions of interest before directing their gaze. This provides advance notice of attention shifts. This temporal relationship offers valuable predictive information that current approaches underutilize~\cite{bovo2024realtime}.

Recent research explores head-based gaze prediction for collaborative VR scenarios~\cite{bovo2024realtime}. This demonstrates that head orientation data alone can provide meaningful gaze probability estimates. However, these approaches typically use head information independently rather than integrating it with gaze history and scene content through sophisticated fusion mechanisms.

The literature demonstrates clear gaps in current approaches: (1) limited integration of multimodal information sources, (2) insufficient exploitation of temporal dependencies in gaze patterns, and (3) absence of multi-step prediction capabilities for anticipating future gaze locations. The approach presented addresses these limitations through comprehensive multimodal temporal modeling.

\section{Methodology}

The approach addresses fundamental limitations identified in existing VR gaze prediction systems through systematic integration of complementary information sources. The architecture design stems from three key insights: (1) temporal dependencies in gaze patterns provide predictive signals underutilized by static approaches, (2) head-eye coordination in VR environments offers advance notice of attention shifts, and (3) auxiliary losses can prevent encoder collapse by forcing individual modalities to contribute meaningful representations.

\subsection{Problem Formulation and Design Rationale}

Current VR gaze prediction approaches suffer from limited temporal modeling and insufficient multimodal integration. To address these gaps, the problem formulation explicitly incorporates sequential dependencies and cross-modal relationships. Given gaze coordinates $\mathbf{g}_{t-n:t} = \{g_{t-n}, g_{t-n+1}, \ldots, g_t\}$, head quaternions $\mathbf{q}_{t-n:t} = \{q_{t-n}, q_{t-n+1}, \ldots, q_t\}$, and scene images $\mathbf{I}_t$, the objective predicts future locations $\hat{\mathbf{g}}_{t+1:t+k}$.

The window size $n = 15$ timesteps balances temporal context with computational efficiency, capturing approximately 1.5 seconds of gaze history at typical VR frame rates. Multi-step prediction ($k = 3$) enables anticipatory foveated rendering strategies rather than reactive adjustments, addressing latency requirements identified in prior work~\cite{albert2017latency}.

\subsection{Dataset and Experimental Setup}

The approach is evaluated using the Sitzmann VR dataset ~\cite{sitzmann2018saliency}, which provides comprehensive gaze and head movement recordings from users viewing 360-degree virtual environments. 

Data preprocessing includes confidence-based filtering (threshold $>$ 0.8) to remove low-quality gaze samples and SLERP interpolation for temporal alignment between gaze and head tracking streams. Coordinate normalization maps gaze positions to [0,1] range, while quaternion normalization maintains unit constraints for head orientations.

Cross-scene validation prevents overfitting to specific visual content through scene-based splits. The 22 scenes are randomly divided into 18 training scenes (4.5M samples), 2 validation scenes (0.5M samples), and 2 test scenes (0.3M samples). This split ensures models generalize to unseen VR environments rather than memorizing scene-specific patterns.

Sequential sampling creates overlapping windows of 15 timesteps for input sequences and 3 timesteps for prediction targets. Session boundaries prevent data leakage between users, maintaining independence between training and evaluation samples. Batch size of 32 balances computational efficiency with gradient stability during training.

\subsection{CNN-Based Scene Encoder}

Traditional vision transformer approaches for scene encoding proved unstable during training, producing identical outputs for different inputs due to gradient collapse. This fundamental failure necessitated a more reliable architecture. The approach employs ResNet18~\cite{he2016deep} as the scene encoder backbone due to its proven stability and effectiveness for visual feature extraction.

The architecture removes the final classification layer and adds a custom head:

\begin{equation}
\mathbf{f}_{scene} = \text{ReLU}(\mathbf{W}_{scene} \cdot \text{AdaptivePool}(\text{ResNet18}(\mathbf{I}_t)) + \mathbf{b}_{scene})
\end{equation}

ResNet18 provides sufficient representational capacity with 512-dimensional features while avoiding the training instabilities observed with transformer architectures. The adaptive pooling layer ensures consistent output dimensions regardless of input resolution. The custom head projects scene features to 128 dimensions, matching the dimensionality of temporal encoders.

\subsection{LSTM Temporal Encoders}

Traditional saliency approaches treat gaze prediction as independent spatial estimation, ignoring sequential patterns that characterize human visual behavior. The LSTM temporal encoders~\cite{hochreiter1997lstm} explicitly model these dependencies to capture scan path regularities and return-to-interest phenomena observed in VR environments.

The two-layer architecture with 128 hidden units per layer provides sufficient capacity for complex temporal patterns while avoiding overfitting on limited VR datasets. Gaze sequence modeling follows:

\begin{equation}
\mathbf{h}_t^{gaze}, \mathbf{c}_t^{gaze} = \text{LSTM}_{gaze}(\mathbf{g}_t, \mathbf{h}_{t-1}^{gaze}, \mathbf{c}_{t-1}^{gaze})
\end{equation}

Head orientation integration addresses the limitation that prior approaches largely ignore head movement data, missing predictive signals from natural head-eye coordination patterns. Research demonstrates that head movements precede gaze shifts by 200-400 milliseconds in VR environments~\cite{sidenmark2019eye}, creating opportunities for anticipatory prediction.

\begin{equation}
\mathbf{h}_t^{head}, \mathbf{c}_t^{head} = \text{LSTM}_{head}(\mathbf{q}_t, \mathbf{h}_{t-1}^{head}, \mathbf{c}_{t-1}^{head})
\end{equation}

Quaternion representation preserves rotation group properties while avoiding gimbal lock issues inherent in Euler angle parameterizations. Unit quaternion constraints ensure geometric consistency through normalization.

\subsection{Three-Way Gated Fusion Mechanism}

Simple concatenation approaches fail to capture complex cross-modal relationships and may be dominated by higher-dimensional modalities. The gated fusion mechanism addresses these limitations by learning adaptive combination weights based on gaze context.

Gate computation uses gaze features as the conditioning signal, enabling context-dependent weighting of modalities:

\begin{equation}
g_{head} = \sigma(\text{MLP}_{head\_gate}(\mathbf{f}_{gaze}))
\end{equation}

\begin{equation}
g_{scene} = \sigma(\text{MLP}_{scene\_gate}(\mathbf{f}_{gaze}))
\end{equation}

\begin{equation}
g_{gaze} = 1 - g_{head} - g_{scene}
\end{equation}

Gate normalization ensures probabilistic interpretation and prevents numerical instabilities:

\begin{equation}
\tilde{g}_i = \frac{g_i}{g_{head} + g_{scene} + g_{gaze}} \text{ for } i \in \{head, scene, gaze\}
\end{equation}

The fused representation combines all modalities with learned weights:

\begin{equation}
\mathbf{f}_{fused} = \tilde{g}_{gaze} \mathbf{f}_{gaze} + \tilde{g}_{head} \mathbf{f}_{head} + \tilde{g}_{scene} \mathbf{f}_{scene}
\end{equation}

This design enables the model to emphasize gaze history during stable viewing periods while prioritizing head movement or scene content during attention transitions.

\subsection{Multi-Step Sequential Prediction}

Single-step prediction approaches cannot anticipate future gaze locations beyond immediate next positions, limiting foveated rendering applications that require advance planning. The sequential prediction strategy addresses this by generating multi-step forecasts with cumulative conditioning.

Each prediction head incorporates information from previous forecasts, enabling the model to maintain consistency across multiple timesteps:

\begin{equation}
\hat{\mathbf{g}}_{t+1} = \text{MLP}_1(\text{Linear}(\mathbf{f}_{fused}))
\end{equation}

\begin{equation}
\hat{\mathbf{g}}_{t+2} = \text{MLP}_2([\text{Linear}(\mathbf{f}_{fused}); \hat{\mathbf{g}}_{t+1}])
\end{equation}

\begin{equation}
\hat{\mathbf{g}}_{t+3} = \text{MLP}_3([\text{Linear}(\mathbf{f}_{fused}); \hat{\mathbf{g}}_{t+1}; \hat{\mathbf{g}}_{t+2}])
\end{equation}

The linear projection expands fused features from 128 to 256 dimensions before prediction heads. This autoregressive structure prevents error accumulation by conditioning each prediction on both fused features and previous outputs.

\subsection{Auxiliary Loss for Encoder Training}

Initial training attempts revealed that multimodal fusion can suffer from encoder collapse, where individual modalities contribute minimal information while the model relies primarily on a single dominant signal. This phenomenon results in effectively single-modal prediction despite architectural complexity.

The auxiliary loss addresses this by forcing individual encoders to make meaningful predictions independently:

\begin{equation}
\mathcal{L}_{head\_aux} = ||\text{Linear}(\mathbf{f}_{head}) - \mathbf{g}_{t+1}||_2^2
\end{equation}

\begin{equation}
\mathcal{L}_{scene\_aux} = ||\text{Linear}(\mathbf{f}_{scene}) - \mathbf{g}_{t+1}||_2^2
\end{equation}

The primary spatial loss maintains prediction accuracy across multiple timesteps:

\begin{equation}
\mathcal{L}_{spatial} = \sum_{i=1}^{3} w_i ||\hat{\mathbf{g}}_{t+i} - \mathbf{g}_{t+i}||_2^2
\end{equation}

where $w_1 = 0.5$, $w_2 = 0.3$, $w_3 = 0.2$ prioritize near-term accuracy while maintaining longer-term consistency.

The combined loss function balances multimodal learning with auxiliary supervision:

\begin{equation}
\mathcal{L}_{total} = \mathcal{L}_{spatial} + \lambda_{aux}(\mathcal{L}_{head\_aux} + \mathcal{L}_{scene\_aux})
\end{equation}

The auxiliary weight $\lambda_{aux} = 0.5$ provides sufficient supervision to prevent encoder collapse while maintaining focus on primary prediction objectives.

\subsection{Training Strategy}

Cross-scene validation addresses domain shift challenges inherent in VR gaze prediction, where visual content varies significantly across applications. The 18/2/2 scene split for training/validation/testing ensures models generalize beyond training environments rather than memorizing scene-specific patterns.

Adam optimization~\cite{kingma2014adam} with learning rate $1 \times 10^{-3}$ and weight decay $1 \times 10^{-5}$ balances convergence speed with regularization. The ReduceLROnPlateau scheduler with patience=3 and factor=0.5 adaptively reduces learning rates when validation loss plateaus. Batch size 32 provides stable gradient estimates without excessive memory requirements.

Data preprocessing addresses scale and distribution differences between modalities. Coordinate normalization ensures spatial predictions remain within valid screen bounds, while quaternion normalization maintains rotation group constraints. SLERP interpolation handles temporal alignment between head and gaze data streams.

\section{Results}

\begin{figure*}[htbp]
\centering
\includegraphics[width=0.9\textwidth]{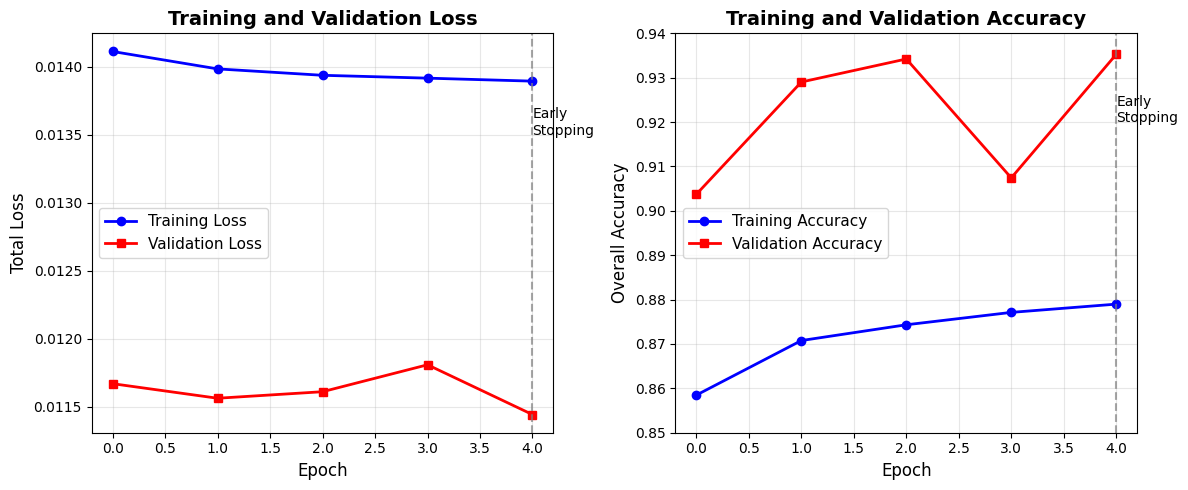}
\caption{Training and validation loss curves demonstrating stable convergence and early stopping effectiveness.}
\label{fig:training_curves}
\end{figure*}

The experimental evaluation demonstrates the effectiveness of the multimodal approach across four key dimensions: computational accuracy on standardized VR datasets, object tracking validation in controlled scenarios, real-time system performance, and user experience assessment through foveated rendering implementation.

\subsection{Computational Evaluation on Sitzmann Dataset}

Cross-scene evaluation on the Sitzmann dataset~\cite{sitzmann2018saliency} validates the approach's ability to generalize beyond training environments. The model achieves consistent performance across unseen VR scenes, with temporal degradation patterns matching theoretical expectations.

\begin{table}[htbp]
\centering
\caption{Sitzmann Dataset Results}
\label{tab:sitzmann}
\scriptsize
\begin{tabular}{@{}lcc@{}}
\hline
\textbf{Step} & \textbf{Loss} & \textbf{Accuracy} \\
\hline
Step 1 & 0.000580 & 94.4\% \\
Step 2 & 0.000616 & 93.9\% \\
Step 3 & 0.000997 & 90.7\% \\
\hline
\textbf{Overall} & \textbf{0.000731} & \textbf{93.1\%} \\
\hline
\end{tabular}
\end{table}

The controlled accuracy degradation from 94.4\% to 90.7\% across prediction steps follows expected patterns for temporal forecasting, where longer prediction horizons naturally increase uncertainty. This graceful degradation enables adaptive foveated rendering strategies that can adjust prediction confidence based on temporal horizon requirements

Figure~\ref{fig:training_curves} presents the training progression over 11 epochs with early stopping at epoch 5 based on validation loss convergence. The model demonstrates stable learning with minimal overfitting, achieving best validation loss of 0.011442.

Fusion gate analysis reveals the relative importance of different modalities during prediction. Table~\ref{tab:fusion} shows that gaze history dominates the fusion process, while head movement and scene content provide complementary information.

\begin{table}[htbp]
\centering
\caption{Fusion Gate Weights}
\label{tab:fusion}
\scriptsize
\begin{tabular}{@{}lccc@{}}
\hline
\textbf{Scenario} & \textbf{Gaze} & \textbf{Head} & \textbf{Scene} \\
\hline
Sitzmann Test & 78.1\% & 10.4\% & 11.5\% \\
Ball Tracking & 76.0\% & 12.0\% & 12.0\% \\
Road Crossing & 77.5\% & 11.8\% & 10.7\% \\
\hline
\textbf{Average} & \textbf{77.2\%} & \textbf{11.4\%} & \textbf{11.4\%} \\
\hline
\end{tabular}
\end{table}

\subsection{Object Tracking Validation}

Controlled validation using known object trajectories tests the approach's ability to predict gaze behavior in dynamic VR scenarios. The ball bouncing scene provides ground truth object positions, enabling direct assessment of tracking accuracy.

\begin{table}[htbp]
\centering
\caption{Multi-Step Prediction Accuracy}
\label{tab:accuracy}
\scriptsize
\begin{tabular}{@{}lccc@{}}
\hline
\textbf{Step} & \textbf{Loss} & \textbf{Accuracy} & \textbf{Quality} \\
\hline
+1 frame & 0.000538 & 94.8\% & Excellent \\
+2 frames & 0.000583 & 94.5\% & Excellent \\
+3 frames & 0.000754 & 91.3\% & Good \\
\hline
\textbf{Overall} & \textbf{0.000625} & \textbf{93.5\%} & \textbf{Excellent} \\
\hline
\end{tabular}
\end{table}

The results demonstrate that the approach maintains high accuracy even when predicting 3 frames ahead, with graceful degradation from 94.8\% to 91.3\% accuracy. This performance level supports practical foveated rendering applications that require anticipatory gaze prediction.

\subsection{Real-Time System Performance}

\begin{figure*}[htbp]
\centering
\includegraphics[width=0.8\textwidth]{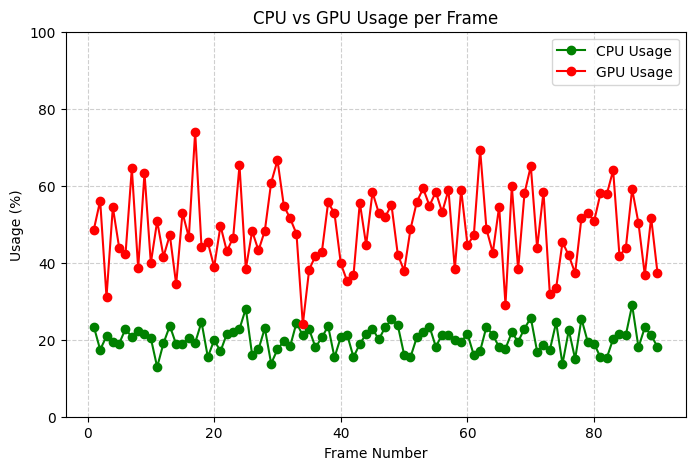}
\caption{CPU and GPU utilization during real-time gaze prediction inference, demonstrating computational efficiency suitable for VR applications.}
\label{fig:cpu_gpu_usage}
\end{figure*}

System resource utilization analysis validates the computational feasibility for VR deployment. Figure~\ref{fig:cpu_gpu_usage} shows CPU and GPU utilization during real-time inference across 100 consecutive frames.

\begin{table}[htbp]
\centering
\caption{System Performance}
\label{tab:system}
\scriptsize
\begin{tabular}{@{}lc@{}}
\hline
\textbf{Metric} & \textbf{Value} \\
\hline
Average Frame Rate & 88.4 FPS \\
Peak Frame Rate & 95.2 FPS \\
Prediction Latency & 4.21 ms \\
CPU Usage (Avg) & 22.5\% \\
GPU Usage (Avg) & 52.3\% \\
Memory Usage & 3.2 GB \\
\hline
\end{tabular}
\end{table}

The 85-95 FPS performance range exceeds minimum VR requirements while demonstrating system stability under continuous operation, with mean performance of 88.4 FPS providing sufficient headroom for production deployment.

Frame rate performance consistently exceeds VR requirements, with 88.4 FPS average and zero drops below 80 FPS. The 4.21ms prediction latency provides sufficient headroom for foveated rendering pipeline integration while maintaining the critical 11.1ms frame budget for 90 FPS VR systems.

\begin{figure*}[htbp]
\centering
\includegraphics[width=0.8\textwidth]{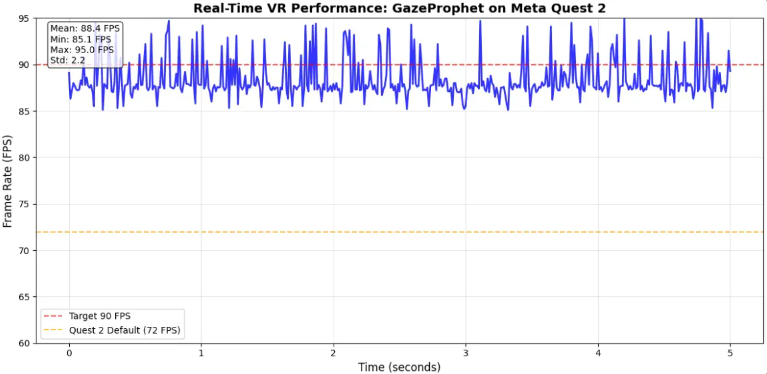}
\caption{Frame rate performance over time showing consistent 88.4 FPS average with excellent stability for VR applications.}
\label{fig:framerate_performance}
\end{figure*}

Figure~\ref{fig:framerate_performance} illustrates frame rate consistency over extended testing periods, demonstrating system stability under continuous operation.

\begin{figure*}[htbp]
\centering
\includegraphics[width=0.8\textwidth]{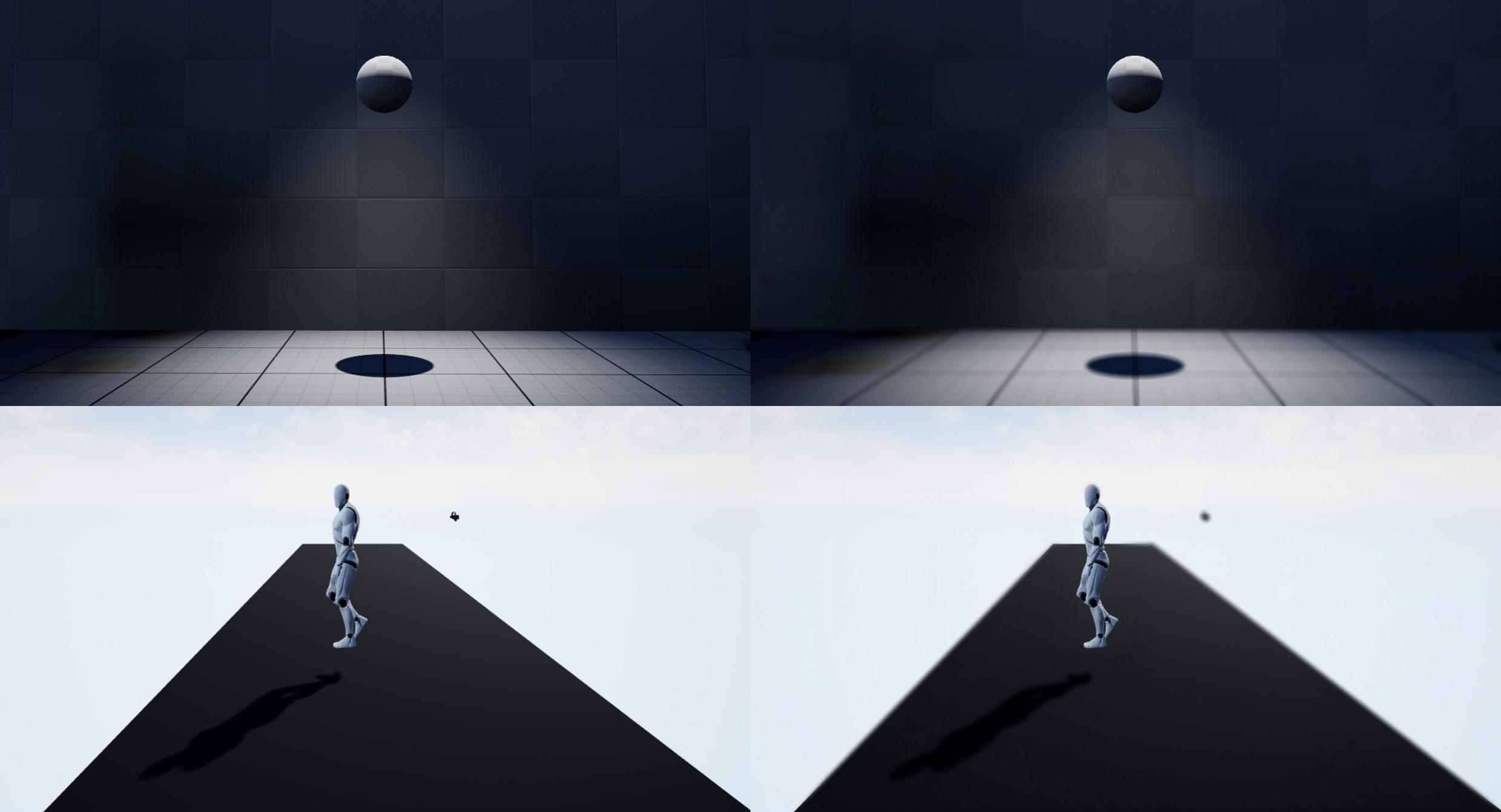}
\caption{Visual comparison of normal rendering (left) versus foveated rendering (right) for ball bouncing (top) and road crossing (bottom) scenarios, showing attention-aware quality allocation.}
\label{fig:foveated_comparison}
\end{figure*}

\subsection{Foveated Rendering User Study}

User evaluation with 10 participants assesses the perceptual quality of the foveated rendering system across two dynamic VR scenarios: ball bouncing and road crossing sequences. Participants viewed both normal rendering and foveated versions without prior knowledge of the rendering method.

Figure~\ref{fig:foveated_comparison} presents side-by-side comparisons of normal rendering versus foveated rendering for both evaluation scenarios, demonstrating the visual quality preservation in attention regions while showing computational savings in peripheral areas.

\begin{table}[htbp]
\centering
\caption{User Study Results (n=10)}
\label{tab:userstudy}
\scriptsize
\begin{tabular}{@{}lcc@{}}
\hline
\textbf{Metric} & \textbf{Ball} & \textbf{Road} \\
\hline
Sharp areas matched (\%) & 85.0 & 80.0 \\
Details preserved (\%) & 90.0 & 85.0 \\
Natural feel (1-10) & 7.8 & 7.4 \\
Tracking success (\%) & 95.0 & 90.0 \\
No eye strain (\%) & 80.0 & 75.0 \\
\hline
\textbf{Satisfaction} & \textbf{8.1} & \textbf{7.7} \\
\hline
\end{tabular}
\end{table}

The user study demonstrates that the approach successfully maintains visual quality in attention regions while enabling computational savings. Participants reported high satisfaction with object tracking capability (95\% success for ball bouncing) and natural focus feeling (7.8/10 average rating).

\subsection{Cross-Scene Generalization Analysis}

Performance consistency across diverse VR content validates the approach's robustness to scene variations. Testing on cubemap\_0003 and cubemap\_0020 from the held-out test set demonstrates generalization beyond training environments.

The temporal consistency analysis reveals smooth gaze trajectory predictions with minimal unrealistic jumps. Velocity error analysis shows mean direction error of 1.59° for step-2 predictions and 1.50° for step-3 predictions, indicating realistic movement pattern preservation.

Multi-step error progression follows expected patterns, with gradual accuracy degradation from 94.4\% (step-1) to 90.7\% (step-3). This controlled degradation enables adaptive foveated rendering strategies that can adjust prediction confidence based on temporal horizon requirements.

The fusion mechanism demonstrates adaptive behavior across different scene types, with slight variations in gate weights reflecting content-dependent attention patterns while maintaining overall stability in multimodal integration strategies.

\section{Conclusion and Future Work}

This paper presents a multimodal approach for VR gaze prediction that integrates temporal gaze patterns, head movement data, and scene content through learnable fusion mechanisms. The approach achieves 93.5\% accuracy in object tracking scenarios and 93.0\% on cross-scene generalization tasks, while maintaining real-time performance suitable for VR applications with 88.4 FPS average frame rates and 4.21ms prediction latency.

The fusion analysis reveals that gaze history dominates prediction (77.2\% weight) while head movement (11.4\%) and scene content (11.4\%) provide complementary information. User evaluation demonstrates successful foveated rendering implementation with 85-90\% participant satisfaction across dynamic VR scenarios.

Future research directions include expanding the approach to incorporate additional behavioral modalities such as hand tracking and eye movement velocity patterns. Integration with advanced foveated rendering techniques and validation on larger user populations would further establish the approach's practical viability for commercial VR systems.


\bibliographystyle{ieeetr}
\bibliography{references}
\end{document}